\documentclass[runningheads,a4paper]{llncs}

\usepackage[mobile]{eccv}
\usepackage{eccvabbrv}
\usepackage{graphicx}
\usepackage[percent]{overpic}
\usepackage{booktabs}
\usepackage{multirow}
\usepackage{amsmath}
\usepackage{amssymb}
\usepackage{enumitem}
\usepackage{float}
\usepackage{colortbl}
\usepackage{capt-of}
\usepackage[most]{tcolorbox}
\usepackage{xcolor}
\usepackage{listings}

\definecolor{toolframe}{HTML}{5A6582}
\definecolor{toolbg}{HTML}{EAECEF}
\definecolor{annotationblue}{RGB}{208,229,245}

\newcommand{\inc}[1]{\ensuremath{_{\text{\textcolor{PineGreen}{(+#1)}}}}}
\newcommand{\dec}[1]{\ensuremath{_{\text{\textcolor{RedOrange}{(-#1)}}}}}

\lstdefinelanguage{json}{
    basicstyle=\ttfamily\scriptsize,
    breaklines=true,
    breakatwhitespace=true,
    frame=none,
    extendedchars=true,
    showstringspaces=false,
    keywordstyle=\bfseries,
    commentstyle=\color{codeigray}
}

\newtcolorbox{cvprbox}[1]{
    enhanced,
    title={#1},
    colframe=toolframe,
    colback=toolbg,
    coltitle=white,
    fonttitle=\bfseries\small,
    halign title=center,
    arc=1.5mm,
    boxrule=0.8mm,
    left=2mm, right=2mm, top=2mm, bottom=2mm,
    toptitle=1mm, bottomtitle=1mm,
    fontupper=\footnotesize
}

\usepackage{hyperref}
\hypersetup{
    pdftitle={Evidence-Backed Video Question Answering},
    pdfauthor={Shijie Wang, Honglu Zhou, Ziyang Wang, Ran Xu, Caiming Xiong, Silvio Savarese, Chen Sun, Juan Carlos Niebles}
}

\begin{document}

\title{Evidence-Backed Video Question Answering}

\author{Shijie Wang\inst{1,2} \and
Honglu Zhou\inst{1} \and
Ziyang Wang\inst{1} \and
Ran Xu\inst{1} \and
Caiming Xiong\inst{1} \and
Silvio Savarese\inst{1} \and
Chen Sun\inst{2} \and
Juan Carlos Niebles\inst{1}}

\authorrunning{S.~Wang et al.}

\institute{Salesforce, Palo Alto, CA, USA
\and
Brown University, Providence, RI, USA}

\maketitle

\begin{abstract}
Current Video Large Language Models (Video LLMs) excel in 
question answering (QA) but largely operate as black boxes, providing textual answers without verifiable visual grounding. Existing explainability efforts rely on textual rationales or sparse bounding boxes, which struggle to capture complex video dynamics such as occlusions and non-rigid deformations. We propose \textbf{Evidence-Backed Video Question Answering (E-VQA)}, a novel task requiring models to jointly output a semantic answer and precise spatio-temporal evidence: temporal segments and dense, tracked object segmentation masklets. To support this, we introduce \textbf{ST-Evidence}, the first human-verified benchmark for both discriminative and generative pixel-level grounding. Evaluations of state-of-the-art models reveal a critical decoupling between QA accuracy and true visual perception that scaling alone fails to bridge. To address this, we develop scalable, automated generation pipelines to create \textbf{ST-Evidence-Instruct}, a 160k-scale dataset bridging high-level reasoning with fine-grained grounding. Fine-tuning grounded Video LLMs on this data yields substantial gains over the corresponding size-matched UniPixel baselines (e.g., +27.2 t-mean and +13.8 $\mathcal{J}$\&$\mathcal{F}$ on a 7B model), establishing a robust baseline for explainable, evidence-backed video understanding. Code and data are available at \url{https://github.com/SalesforceAIResearch/EVQA}.

\end{abstract}
    
\section{Introduction}
\label{sec:intro}
Recent Video Large Language Models (Video LLMs) have demonstrated impressive results on Video Question Answering (Video QA) benchmarks. However, most remain black-box systems that provide only textual answers, raising significant concerns regarding trust, explainability, and reliability. Without verifiable evidence, these models may rely on language priors or hallucinations rather than genuine visual perception; when errors occur, tracing their origin is nearly impossible. Although recent work explores video reasoning via language-based chain-of-thought (CoT) reasoning~\cite{wei2022chain,feng2025video,li2025videochatr1,wang2025video,han2025videoespresso}, these textual rationales often lack grounding in concrete spatio-temporal visual evidence. This lack of traceability is particularly critical in high-stakes domains such as autonomous driving, medical procedure analysis and human--robot collaboration, where decision-making requires rigorous visual justification.

To bridge the gap between semantic reasoning and visual perception, we argue that effective video understanding requires evidence grounding: the ability to explicitly justify an answer with its corresponding spatio-temporal visual evidence. Existing grounded video reasoning work~\cite{cheng2025vstarbenchmarkingvideollmsvideo, yan2024visa, meng2025open} predominantly focuses on answer grounding (identifying the object that constitutes the answer), often relying on spatially sparse bounding boxes over temporally limited keyframes. However, such sparse grounding lacks the precision to capture complex video dynamics, such as severe occlusions, continuous state changes, non-rigid deformations (e.g., liquids, wires), or identity preservation through crossovers (e.g., tracking a specific cup in a shell game). We argue that to unambiguously verify that a model has perceived the correct visual cues, it must perform dense, pixel-level spatio-temporal tracking as a formal evidence trail.

\begin{figure}[t]
    \centering
    \includegraphics[width=\linewidth]{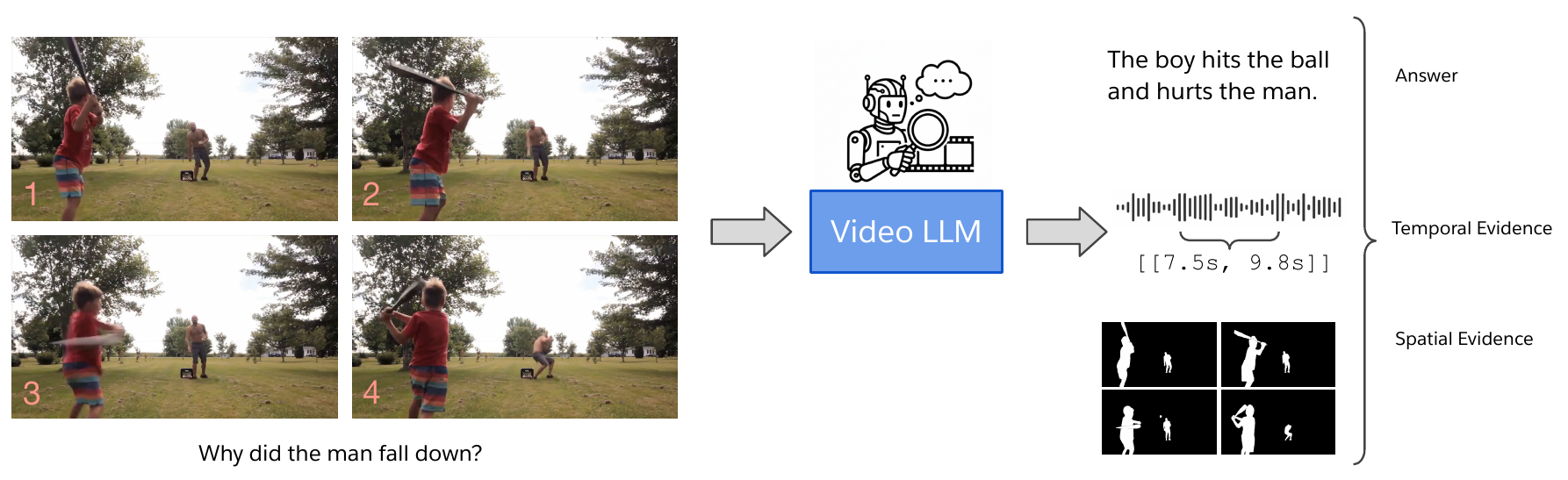}
      \vspace{-15pt}

    \captionof{figure}{\textbf{E-VQA: Evidence-Backed Video Question Answering}. Models provide textual answers to video questions while grounding their reasoning in spatio-temporal evidence, including relevant temporal video segments and densely tracked segmentation masks that highlight the spatio-temporal visual regions supporting the answer.}
    \label{fig:teaser}
        \vspace{-10pt}

\end{figure}

To formalize this task, we introduce \textbf{Evidence-Backed Video Question Answering} (\textbf{E-VQA}) (Fig.~\ref{fig:teaser}). Given a video and a question, a model must generate a triplet: (i) the semantic answer, (ii) the supporting temporal evidence (relevant video segments), and (iii) the supporting spatial evidence represented as dense, tracked spatio-temporal segmentation masks (masklets). By requiring these components, E-VQA explicitly couples high-level reasoning with low-level grounding. This formulation forces a model to not only solve the linguistic task but also to provide a verifiable, pixel-level justification of its perception.

As no existing datasets support this setting, we introduce \textbf{ST-Evidence}, the first benchmark specifically designed for E-VQA. ST-Evidence is constructed through a rigorous three-stage semi-automatic pipeline:
\textit{(1) Sample}: Select
 suitable video–question–answer pairs from existing Video QA datasets;
\textit{(2) Annotate}: Human annotators provide dense spatio-temporal evidence or reject low-quality samples;
and \textit{(3) Verify}: Independent human reviewers assess annotation quality and flag samples for re-annotation if necessary.

Since current General Video LLMs are typically not designed for dense, generative spatio-temporal grounding, we propose two benchmark variants:

\noindent\textbf{ST-Evidence-Gen}: A generative setting requiring the model to synthesize the complete triplet, coupling reasoning with dense pixel-level grounding.

\noindent\textbf{ST-Evidence-MCQ}: A multiple-choice formulation where the model must (1) select the correct answer, (2) identify the relevant temporal segment, and (3) choose the correct spatial mask from a candidate set. This variant focuses on discriminative and sparse grounding.

We evaluate leading General Video LLMs—including open-source models such as Qwen3-VL~\cite{qwen2025qwen3vl} and proprietary systems such as OpenAI-o3~\cite{OpenAI_o3_2024} and Gemini-2.5-Pro~\cite{comanici2025gemini25pushingfrontier}.
Our results on ST-Evidence-MCQ show that high QA accuracy does not correlate with grounding proficiency;
most open-source models perform near random-guess levels in spatial evidence selection.
For ST-Evidence-Gen, we employ a two-step pipeline where General Video LLMs answer the question, generate temporal segments, and describe key evidence objects in text, which are then parsed by a multimodal grounding model (UniPixel~\cite{liu2025unipixel}) to produce spatio-temporal masks.
Under this regime, both temporal and spatial grounding scores remain low.
Even specialized \textit{Grounded} Video LLMs, such as UniPixel~\cite{liu2025unipixel} and Sa2VA~\cite{yuan2025sa2va}, struggle with the joint reasoning required for E-VQA.
Our comprehensive evaluations suggest that scaling alone is insufficient; progress in E-VQA demands fundamental advances in architecture, data quality, and integrated training objectives.

To address these limitations, we introduce \textbf{ST-Evidence-Instruct}, a large-scale instruction-tuning dataset containing \textbf{160k} triplets of (QA pairs, temporal evidence, spatial evidence). While prior datasets are either purely \textit{semantic} (QA) or \textit{grounded} (localization), we bridge this gap by synthesizing aligned spatio-temporal evidence for existing QA pairs and by repurposing masklet-annotated videos for reasoning tasks. Our fully automatic pipelines enable generation of the required triplets at scale. Fine-tuning UniPixel~\cite{liu2025unipixel} on \textbf{ST-Evidence-Instruct}
yields improvements in QA accuracy and significant gains in dense grounding, validating our data-centric approach to enhancing E-VQA performance.

Our main contributions are summarized as follows:
\textbf{(1) A new task:} We introduce Evidence-Backed Video Question Answering (\textbf{E-VQA}), a task requiring models to provide verifiable justification through the joint output of semantic answers, temporal segments, and dense spatial masklets. 
\textbf{(2) A comprehensive benchmark:} We construct ST-Evidence, the first E-VQA benchmark, featuring human-verified spatio-temporal annotations across generative (ST-Evidence-Gen) and discriminative (ST-Evidence-MCQ) evaluation variants.
\textbf{(3) Large-scale instruction-tuning dataset:} We develop scalable, automated pipelines to bridge semantic and grounded video data, yielding ST-Evidence-Instruct with 160k triplets. Fine-tuning current grounded Video LLMs on this dataset achieves considerable gains in joint reasoning and pixel-level evidence grounding.

\section{Related Work}
\label{sec:related}

\noindent\emph{Video Large Language Models.}
Video LLMs extend LLMs to reason over multimodal video inputs. While proprietary (e.g., GPT-4o~\cite{openai_gpt4o} and o3~\cite{OpenAI_o3_2024}, Gemini 2.5~\cite{comanici2025gemini25pushingfrontier}) and open-source models (e.g., Video-LLaMA3~\cite{damonlpsg2025videollama3}, InternVL-3.5~\cite{wang2025internvl3}, Qwen3-VL~\cite{qwen2025qwen3vl}) show strong comprehension, they generate purely textual responses without visual justification.
To improve logical depth, recent Reasoning Video LLMs apply reinforcement learning (RL) and test-time scaling to generate multi-step reasoning chains (e.g., Video-R1~\cite{feng2025video}, VideoChat-R1~\cite{li2025videochatr1}, Video-RTS~\cite{wang2025video}). Some emerging models attempt to ground these chains: VITED~\cite{lu2025vited} and Time-R1~\cite{wang2025time} focus on temporal segments, while Open-O3 Video~\cite{meng2025open}, Seg-R1~\cite{you2025seg} and others~\cite{gong2025reinforcing} incorporate sparse bounding boxes or ``think-before-segment'' strategies. 
However, these works typically treat grounding as an intermediate ``scratchpad'' to improve textual accuracy. In contrast, E-VQA requires dense, spatio-temporal masklets as a formal component of the final answer triplet, demanding a pixel-level ``proof'' of the model's underlying reasoning.

\noindent\emph{Spatio-Temporal Grounded Video LLMs.}
Efforts to bridge language and perception have yielded Temporally Grounded Video LLMs (VTimeLLM~\cite{huang2024vtimellm}, Momentor~\cite{qian2024momentor}, VTG-LLM~\cite{guo2024vtg}) that predict segment boundaries but lack spatial grounding. 
To add spatial grounding, some models (VideoMolmo~\cite{ahmad2025videomolmo}, NumPro~\cite{wu2025number}, VGR~\cite{wang2025vgr}) utilize bounding boxes,
while pixel-level models (Sa2VA~\cite{yuan2025sa2va}, UniPixel~\cite{liu2025unipixel}, VideoLISA~\cite{bai2024one}, VideoGLaMM~\cite{munasinghe2024videoglamm}, GLUS~\cite{lin2025glus}) integrate segmentation decoders for dense masks.
However, these methods often treat grounding and QA as separate tasks,
rather than addressing reasoning and justification jointly.
We address this via instruction tuning that compels models to synergize high-level reasoning with dense spatio-temporal tracking, moving beyond disjoint task execution.

\noindent\emph{Video QA and Grounding Datasets.}
Existing datasets
do not fully support dense, evidence-backed reasoning. Semantic datasets, such as NeXT-QA~\cite{xiao2021next} and STAR~\cite{wu2021star_situated_reasoning}, focus on causal reasoning but lack grounding annotations. Conversely, recent grounded video datasets have emerged, such as NeXT-GQA~\cite{xiao2024can}, V-STaR~\cite{cheng2025vstarbenchmarkingvideollmsvideo}, VideoEspresso~\cite{han2025videoespresso}, and CG-Bench~\cite{chen2024cg}, as well as referential comprehension video datasets (e.g., SAMA~\cite{sun2025sama}, VideoRefer~\cite{yuan2025videorefer}, and Strefer~\cite{zhou2025strefer}).
Earlier grounded Video QA efforts expose supporting evidence in different forms: STAIR~\cite{wang2024stair} audits intermediate spatio-temporal results, TranSTR~\cite{li2023transtr} selects question-critical moments and objects as rationales, and TVQA+~\cite{lei2020tvqaplus} augments Video QA with temporal moments and sparse bounding boxes. In contrast, E-VQA requires temporal segments and dense tracked masklets as components of the final output rather than as intermediate or sparse rationales.
Domain-specific efforts also include EgoMask~\cite{liang2025fine} and InterRVOS~\cite{jin2025interrvos}.
However, E-VQA differs in two key ways:
(a) Evidence vs. Answer Grounding: while prior work primarily focuses on \textit{answer grounding} (locating the visual entity that \emph{is} the answer), E-VQA requires \textit{evidence grounding}---locating the visual cues that logically lead to the answer.
(b) Dense vs. Sparse Representation: most existing benchmarks rely on sparse annotations (segments or bounding boxes on keyframes). As E-VQA demonstrates, these are insufficient for complex video dynamics such as severe occlusions, continuous state changes, and non-rigid object interactions. Our ST-Evidence benchmark unifies high-level reasoning with dense evidence generation, requiring tracked masklets at 6 FPS, and thus establishes a rigorous standard that binds linguistic logic to verifiable visual facts.

\section{Evidence-Backed Video QA}
\label{sec:benchmark}

We introduce \textbf{Evidence-Backed Video Question Answering (E-VQA)}. Unlike standard Video Question Answering (VQA), which only predicts an answer $A$, E-VQA requires a model $F$ to justify its prediction via a unified triplet $(A, E_{t}, E_{s})$, given a video $V$ and a question $Q$. The supporting spatio-temporal evidence is defined as:

    \noindent \textbf{Temporal Evidence ($E_{t}$):} A set of non-overlapping time segments containing the critical time spans essential to answering $Q$:
    {\scriptsize
    \[
        E_t = \bigl\{[\mathit{start}_i,\mathit{end}_i]\bigr\}_{i=1}^{N}.
    \]}
    
    \noindent \textbf{Spatial Evidence ($E_s$):} The key objects or regions relevant to the reasoning process, represented as spatio-temporal masks (i.e., ``masklets'').

By requiring models to ground their reasoning in specific spatio-temporal regions, E-VQA mitigates reliance on static language priors (i.e., ``language bias'') and significantly enhances the explainability of video understanding models.

\subsection{Benchmark Design: ST-Evidence}
To rigorously evaluate models on the E-VQA task, we introduce \textbf{ST-Evidence}. The benchmark is constructed via a semi-automatic three-stage pipeline---\emph{Sample}, \emph{Annotate}, and \emph{Verify}---ensuring high-quality, human-verified evidence.

  \noindent \textbf{(1) Sample.} We sample high-quality video-question pairs from the validation and test splits of existing semantic Video QA benchmarks, including NeXT-QA \cite{xiao2021next}, Perception Test \cite{patraucean2023perception}, STAR \cite{wu2021star_situated_reasoning}, CLEVRER \cite{CLEVRER2020ICLR}, and Ego4D~\cite{grauman2022ego4d}. As not all samples are suitable, we employ a vision-language model (VLM) as an initial filter. The filtering criteria require: (i) clear video quality and an appropriate duration (5--200~s), (ii) temporal evidence that spans neither the entire video nor a single frame, and (iii) a specific, visually determinable question.
  
  \noindent \textbf{(2) Annotate.} Samples passing the VLM filter are assigned to human experts. Provided with the video, question, and ground-truth answer, annotators are tasked with: (i) identifying all critical temporal evidence segments, and (ii) annotating spatio-temporal tracked segmentation masks for each evidence object. Masks are densely annotated at 6 FPS, a standard sampling rate used in prominent video segmentation datasets (e.g., DAVIS~\cite{pont20172017}) to balance fine-grained temporal resolution with annotation feasibility. Annotators also filter out any remaining ambiguous or subjective samples.
  
  \noindent \textbf{(3) Verify.} Finally, all annotations are reviewed by PhD-level researchers. Any suboptimal annotations are flagged for re-annotation, ensuring the benchmark maintains a rigorous standard for evaluating precise evidence grounding.

Given that dense mask annotation is exceptionally labor-intensive, we supplement our benchmark by sourcing data from the validation split of ViCaS~\cite{athar2024vicas}, a dataset containing human-annotated dense masks and captions. We use a VLM to generate questions, answers, and distractor options conditioned on the video and dense captions. Human annotators then filter unsuitable QA pairs, annotate the temporal evidence, and map the correct spatial evidence to the existing human-annotated ViCaS object masks. 
We use Qwen3-VL-235B-A22B~\cite{qwen2025qwen3vl} as the VLM.

\paragraph{Benchmark Variants.} 
We propose ST-Evidence-Gen for generative, pixel-level grounding and ST-Evidence-MCQ for multiple-choice discriminative grounding. These distinct formats accommodate a broad spectrum of model architectures, evaluating E-VQA through both open-ended synthesis and structured selection.

\textbf{ST-Evidence-Gen} (Generative) requires a model to generate the complete triplet $(A, E_t, E_s)$. This entails selecting the correct answer $A$ from a list of options, predicting temporal segments $E_t$ (start/end timestamps), and producing the spatial evidence $E_s$ (as masklets). We evaluate $A$ via accuracy; $E_t$ using temporal Intersection over Union (tIoU) and Intersection over Prediction (IoP); and $E_s$ via the standard $\mathcal{J}$\&$\mathcal{F}$ metric to jointly consider region similarity $\mathcal{J}$ and contour accuracy $\mathcal{F}$. See the Supplementary Material for details.

\textbf{ST-Evidence-MCQ} (Multiple-Choice Question) is designed for general Video LLMs not explicitly trained for generative localization. In this setting, all three subtasks are formulated as multiple-choice questions, requiring the model to: (1) select the correct answer $A$ from a list of options, (2) select the correct temporal evidence $E_t$ from a set of candidate segments, and (3) select the correct spatial evidence $E_s$ from a set of candidate masks.

For temporal evidence, we use Qwen3-VL-235B-A22B~\cite{qwen2025qwen3vl} to generate plausible yet deliberately incorrect distractors, ensuring they are semantically relevant but have minimal overlap with the ground-truth segments.

For spatial evidence, we first sample the middle frame of a ground-truth evidence segment as the reference frame; the corresponding human-annotated mask serves as the answer. To ensure high quality and keep the task challenging, we then manually annotate three plausible but incorrect object masks as distractor options to formulate four-way multiple-choice questions.

All three subtasks are evaluated using standard accuracy.

\section{Instruction Tuning Data Construction}
\label{sec:training_data}

\vspace{-5pt}

\begin{figure*}[t]
    \centering
    \begin{overpic}[width=1.0\linewidth]{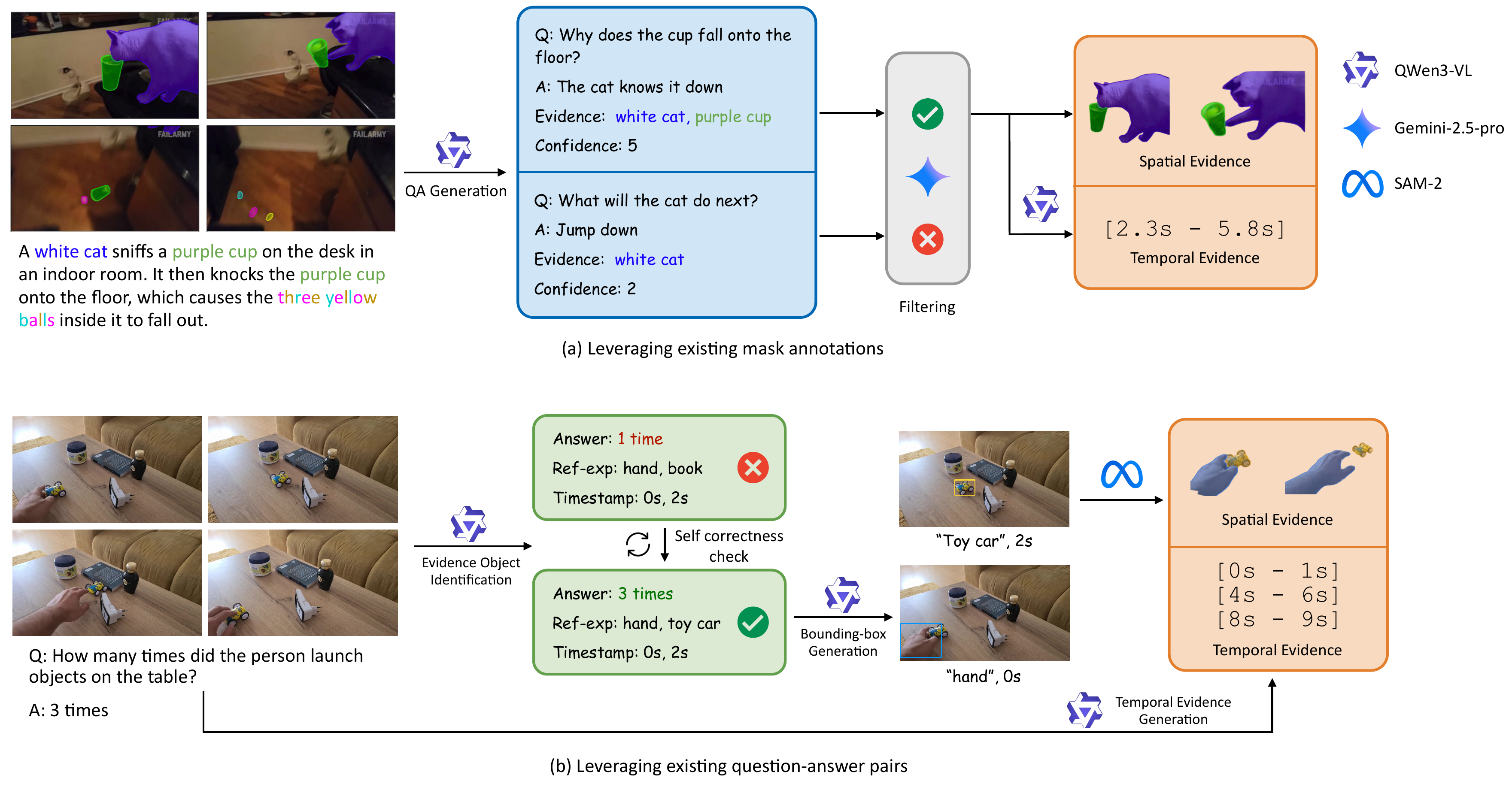}
        \put(35.40,45.55){\begingroup\setlength{\fboxsep}{0.15pt}\colorbox{annotationblue}{\makebox[15.3mm][l]{\resizebox{15.1mm}{1.8mm}{\sffamily A: The cat knocks it down}}}\endgroup}
        \put(44.55,14.50){\begingroup\setlength{\fboxsep}{0pt}\colorbox{white}{\rule{0pt}{3.1mm}\makebox[9.0mm]{}}\endgroup}
        \put(44.70,14.70){\resizebox{8.7mm}{2.7mm}{\sffamily\shortstack{Self-consistency\\check}}}
    \end{overpic}
      \vspace{-20pt}
    \captionof{figure}{Overview of our fully automated pipelines for constructing ST-Evidence-Instruct, a 160k-scale dataset for E-VQA instruction tuning.}
    \label{fig:annotation}
    \vspace{-10pt}
\end{figure*}

Effective spatio-temporal reasoning requires large-scale training data aligning QA with dense grounded annotations. Existing datasets generally fall into two categories: \emph{semantic} and \emph{grounded}.
\emph{Semantic} datasets (e.g., video QA) focus on holistic understanding but lack grounding, while \emph{grounded} datasets (e.g., temporal or spatial localization) target fine-grained perception and localization accuracy, often at the expense of complex reasoning.

The Evidence-Backed Video Question Answering (E-VQA) task unifies these two paradigms, demanding both high-level semantic reasoning and fine-grained spatio-temporal grounding. This dual requirement makes large-scale data annotation from scratch prohibitively challenging and costly. Therefore, to support future research on E\mbox{-}VQA, we propose scalable data-annotation pipelines designed to construct E-VQA training datasets by leveraging both existing semantic and grounded video understanding datasets.

\begin{table*}[tb]
    \caption{Statistics of our proposed datasets for the \textbf{E-VQA} task.}
        \vspace{-10pt}

    \centering
    \small
    \resizebox{\textwidth}{!}{
    \begin{tabular}{cccccc}
        \toprule
         Name & \# Questions & \# Videos & Data sources & Temporal annotation & Mask annotation\\
         \midrule
         ST-Evidence-MCQ & 1298 & 348 & NeXT-QA & Human & Human \\
         ST-Evidence-Gen & 2706 & 2548 & Perception Test, STAR, CLEVRER, ViCaS, Ego4D & Human & Human \\
         ST-Evidence-Instruct & 160k  &30k & Perception Test, STAR, CLEVRER, ViCaS & Model &Model + Human \\

         \bottomrule
    \end{tabular}
      \vspace{-15pt}

    }
    \label{tab:benchmark}
\end{table*}

\noindent\textbf{Data Sources and Statistics.} 
We construct our training dataset by aggregating samples from two benchmark categories: (1) Video Question Answering and (2) Video Segmentation. Specifically, as sources for our pipelines, we sample 20k video-question pairs from the training sets of Perception Test \cite{patraucean2023perception}, STAR \cite{wu2021star_situated_reasoning}, and CLEVRER \cite{CLEVRER2020ICLR}. We also sample 20k videos with detailed captions and segmentation masks from the ViCaS~\cite{athar2024vicas} training set. Table~\ref{tab:benchmark} summarizes the data statistics.

\subsection{Automatic Data Construction Pipeline}

Each sample in the E-VQA dataset is formulated as a tuple $(V, Q, A, C, E_t, E_s)$, comprising a video $V$, a question $Q$, an answer $A$, optional candidate choices $C$, temporal evidence $E_t$ (time segments), and spatial evidence $E_s$ (dense video masklets). To construct this at scale, we design a bidirectional, multi-step generation paradigm based on the source data type (illustrated in Figure~\ref{fig:annotation}).

\noindent \textbf{Pathway 1: Grounding-to-Semantics (Leveraging Existing Masks)}.
For datasets such as ViCaS~\cite{athar2024vicas}, which already contain dense captions and pixel-level object masks paired with natural language object phrases (i.e., referring expressions), we employ a generation-verification pipeline to synthesize grounded QA pairs as shown in Figure~\ref{fig:annotation}(a).

    \textbf{(1) Generation.} ViCaS provides human-written captions with phrase grounding linked to object masks. We utilize two VLMs (Qwen3-VL and Gemini) to independently generate tuples containing: a question $Q$, an answer $A$ with distractor options $C$, the associated evidence objects (specifically, object phrases), a confidence score, and a brief explanation. The models are conditioned on the video, the ground-truth captions, and the list of candidate objects. We employ rigorous prompting strategies (detailed in the Supplementary Material) to ensure the questions are answerable and grounded in the provided objects. For each video, each model generates three to four candidate questions, yielding typically six to eight candidates in total.

    \textbf{(2) Filtering.} We employ a two-stage quality control process. First, we discard samples with formatting errors (e.g., missing answers, hallucinated evidence objects) or low confidence scores ($<\frac{3}{5}$). Second, we perform text-only validation using Gemini\mbox{-}2.5\mbox{-}Pro, which reviews the captions, object candidates, generated QA, evidence objects, and explanation to accept or reject each sample.

    \textbf{(3) Evidence Assignment.} For each accepted video-question pair, we query Qwen3-VL again to predict the temporal evidence $E_t$. Notably, we process videos at a higher frame rate (4 FPS) during this stage than during the question-generation stage (2 FPS) to capture finer temporal dynamics. Finally, the spatial evidence $E_s$ is directly derived from the pre-existing ViCaS mask annotations corresponding to the identified evidence objects.

\noindent \textbf{Pathway 2: Semantics-to-Grounding (Leveraging Existing QA)}
For semantic datasets containing complex QA pairs but lacking visual grounding, the primary challenge lies in generating the supporting evidence ($E_t$ and $E_s$). While temporal evidence $E_t$ is generated using the method described in Pathway 1, generating accurate spatial masks $E_s$ from complex questions remains non-trivial. A significant domain gap exists between \emph{reasoning models} and \emph{grounding models}. Reasoning models (typically trained on VQA) excel at processing complex queries but lack fine-grained spatial output capabilities (i.e., they usually generate pure text output). In contrast, grounding models (typically trained on localization tasks) produce precise pixel-level predictions but are limited to simple referring expressions or visual prompts, failing on complex reasoning tasks. To bridge this gap, we propose a three-step decomposition pipeline that leverages the strengths of both paradigms, as shown in Figure~\ref{fig:annotation}(b):

    \textbf{(1) Evidence Object Identification.} We first employ Qwen3-VL-235B-A22B to identify the specific objects required to answer the question. To ensure quality, we adopt an intuitive principle: ideal evidence objects are those most likely to lead to the correct answer. We prompt the VLM to: (i) list referring expressions that uniquely describe the critical objects, (ii) provide a timestamp at which each object is clearly visible, and (iii) answer the original question based on these objects. If the generated answer matches the ground truth, the evidence is accepted. We allow up to five re-prompting attempts; if the model fails to predict the correct answer within five attempts, the sample is discarded.

    \textbf{(2) Bounding Box Generation.} Given the verified referring expressions and their associated timestamps, we extract the corresponding frames and prompt Qwen3-VL to predict a spatial bounding box for each object. To ensure robust downstream tracking, we apply a filtering mechanism to remove degenerate predictions. Specifically, we discard bounding boxes exhibiting extreme area ratios ($<1\%$ or $>90\%$), anomalous aspect ratios, or severe spatial inconsistencies compared to the object's bounding boxes in adjacent sampled frames.

    \textbf{(3) Dense Mask Propagation.} Finally, we utilize SAM-3~\cite{carion2025sam3segmentconcepts} to propagate dense spatio-temporal segmentation masks across the entire video volume, using the verified multi-frame bounding boxes as prompt inputs. To eliminate redundancy and ensure a concise set of spatial evidence, we apply a post-processing filter that deduplicates highly overlapping masklets (IoU $> 0.9$).

This pipeline yields \textbf{ST-Evidence-Instruct}, a 160k-scale dataset containing aligned $(Q, A, E_t, E_s)$ tuples, providing the necessary data foundation to train models capable of explainable, evidence-backed video reasoning.

\section{Experiments}
\label{sec:experiment}

\begin{table*}[t]
  \caption{Performance on ST-Evidence-Gen. $^{\dagger}$ indicates closed-source proprietary models. For General Video LLMs, spatial evidence is generated via a proxy segmentation model (UniPixel-3B). Fine-tuning gains are shown relative to the size-matched UniPixel baseline (Ours-3B vs. UniPixel-3B; Ours-7B vs. UniPixel-7B).}
    \vspace{-5pt}

\setlength{\tabcolsep}{2.5pt}
\aboverulesep=0ex
\belowrulesep=0ex 
\centering
\footnotesize
\resizebox{\textwidth}{!}{
  \begin{tabular}{l c ccc ccc} 
\toprule
Methods &
QA &
\multicolumn{3}{c}{T-Evidence} &
\multicolumn{3}{c}{S-Evidence} \\

\cmidrule(lr){2-2}\cmidrule(lr){3-5}\cmidrule(lr){6-8} 
 & acc & mIoU & mIoP &t-mean & $\mathcal{J}$ & $\mathcal{F}$ & $\mathcal{J}$\&$\mathcal{F}$ \\
\midrule

\rowcolor{gray!15}\multicolumn{8}{l}{\textit{General Video LLM + Segmentation Model (UniPixel-3B)}} \\
OpenAI-o3$^{\dagger}$~\cite{OpenAI_o3_2024} & 82.6 & 30.6 & 41.6 & 36.1 & 39.4 & 44.4 & 41.9 \\
Gemini-2.5-Flash$^{\dagger}$~\cite{comanici2025gemini25pushingfrontier} & 81.2 & 27.2 & 43.1 & 35.2 & 40.5 & 45.4 & 42.9 \\
Gemini-2.5-Pro$^{\dagger}$~\cite{comanici2025gemini25pushingfrontier}  & \textbf{83.7} & \textbf{42.8}  & \textbf{57.1}  & \textbf{49.9}  & 41.8 & 46.3 & 44.0 \\
Video-LLaMA3-7B~\cite{damonlpsg2025videollama3} &51.0 &13.7 &18.9 &16.3 &12.3 &14.2 &13.2 \\
LLaVA-OV-1.5-8B~\cite{an2025llava} &67.7 &13.6 &26.9 &20.3 &35.5 &40.1 &37.8 \\
InternVL-3.5-8B~\cite{wang2025internvl3} &72.0 &17.5 &27.8 &22.6 &33.7 &39.7 &36.7 \\
Qwen2.5-VL-3B~\cite{bai2025qwen25vltechnicalreport} &72.4 &21.8 &27.4 &24.6 &29.6 &33.5 &31.6 \\
Qwen2.5-VL-7B~\cite{bai2025qwen25vltechnicalreport} &73.4 &18.1 &28.5 &23.3 &27.7 &30.9 &29.3 \\
Qwen2.5-VL-72B~\cite{bai2025qwen25vltechnicalreport} &76.2 &20.5 &30.1 &25.3 &36.6 &42.3 &39.5 \\
Qwen3-VL-4B~\cite{qwen2025qwen3vl} &76.9 &33.0 &44.3 &38.6 &37.6 &42.3 &39.9 \\
Qwen3-VL-8B~\cite{qwen2025qwen3vl} &78.2 &30.5 &43.3 &36.9  &39.0 &43.3 &41.1 \\
Qwen3-VL-235B-A22B~\cite{qwen2025qwen3vl} &\underline{83.6} &\underline{38.4} &\underline{50.6} &\underline{44.5} &39.6 &44.1 &41.9 \\
\midrule

\rowcolor{gray!15}\multicolumn{8}{l}{\textit{Grounded Video LLM}} \\

Sa2VA-Qwen2.5-VL-7B~\cite{yuan2025sa2va} &76.8 &0.0 &0.0 &0.0 &23.5  &28.8  &26.2 \\
UniPixel-3B~\cite{liu2025unipixel} &72.2 &4.5 &8.6 &6.5 &40.2 &45.2 &42.7 \\
UniPixel-7B~\cite{liu2025unipixel} &75.0 &1.4 &2.4 &1.9 &38.3 &42.4 &40.3 \\
\textbf{Ours-3B} &72.4\inc{0.2} &20.6\inc{16.1} &33.2\inc{24.6} &26.9\inc{20.4} &\underline{51.2}\inc{11.0} &\underline{54.3}\inc{9.1} &\underline{52.7}\inc{10.0} \\

\textbf{Ours-7B} &76.0\inc{1.0} &21.1\inc{19.7} &37.1\inc{34.7} &29.1\inc{27.2} &\textbf{52.8}\inc{14.5} &\textbf{55.4}\inc{13.0} &\textbf{54.1}\inc{13.8} \\

\bottomrule
\end{tabular}
}

\label{tab:gen_table}
\end{table*}

\begin{table*}[t]
  \caption{Performance on ST-Evidence-MCQ. $^{\dagger}$ indicates closed-source models.}
    \vspace{-5pt}

\setlength{\tabcolsep}{2.5pt}
\aboverulesep=0ex
\belowrulesep=0ex 
\centering
\footnotesize
\resizebox{\textwidth}{!}{
  \begin{tabular*}{\textwidth}{@{\extracolsep{\fill}}l c c c} 
\toprule
Methods &
QA-acc &
T-Evidence-acc &
S-Evidence-acc \\
\midrule

\textit{Random Guess} & 20.00 & 25.00 & 25.00 \\
\midrule

\rowcolor{gray!15}\multicolumn{4}{l}{\textit{General Video LLM}} \\
OpenAI-o3$^{\dagger}$~\cite{OpenAI_o3_2024} &81.21 &50.67 &\underline{84.32} \\
Gemini-2.5-Flash$^{\dagger}$~\cite{comanici2025gemini25pushingfrontier} & 80.43 & \textbf{78.81} & 36.29 \\
Gemini-2.5-Pro$^{\dagger}$~\cite{comanici2025gemini25pushingfrontier} & \textbf{82.82} & 70.88 & 81.28 \\
Video-LLaMA3-7B~\cite{damonlpsg2025videollama3} &67.64 &48.84 &27.12 \\
LLaVA-OV-1.5-8B~\cite{an2025llava} &69.03 &66.02 &23.42  \\
InternVL-3.5-8B~\cite{wang2025internvl3} & 79.04 &58.17 &27.43 \\
Qwen2.5-VL-3B~\cite{bai2025qwen25vltechnicalreport} & 71.42 & 45.69 & 26.35 \\
Qwen2.5-VL-7B~\cite{bai2025qwen25vltechnicalreport} & 74.65 & 27.12 & 46.15 \\
Qwen2.5-VL-72B~\cite{bai2025qwen25vltechnicalreport} & 79.04 & 45.38 & 71.80 \\
Qwen3-VL-4B~\cite{qwen2025qwen3vl} & 76.44 & 63.41 & 77.04 \\
Qwen3-VL-8B~\cite{qwen2025qwen3vl} & 77.81 & 70.34 & 76.43 \\
Qwen3-VL-30B-A3B~\cite{qwen2025qwen3vl} & 81.14 & \underline{71.54} & 67.04 \\
Qwen3-VL-235B-A22B~\cite{qwen2025qwen3vl} & \underline{81.59} & 70.96 & \textbf{86.90} \\
\midrule

\rowcolor{gray!15}\multicolumn{4}{l}{\textit{Grounded Video LLM}} \\
Sa2VA-Qwen2.5-VL-7B~\cite{yuan2025sa2va} & 62.40 & 34.21 & 23.04 \\
UniPixel-3B~\cite{liu2025unipixel}& 68.57 & 45.45 & 24.88 \\
UniPixel-7B~\cite{liu2025unipixel} & 74.04 & 35.29 & 22.03 \\
\bottomrule
\end{tabular*}}

\label{tab:mcq_table}
\vspace{-5pt}
\end{table*}

\begin{table*}[t]
  \caption{Performance on video segmentation and general understanding benchmarks. Gains or losses are shown relative to the size-matched UniPixel baseline.}
  \vspace{-5pt}

  \setlength{\tabcolsep}{2.5pt}
  \aboverulesep=0ex
  \belowrulesep=0ex
  \centering
  \footnotesize
  \begin{tabular*}{\textwidth}{@{\extracolsep{\fill}}l c ccc c}
    \toprule
    \multirow{2}{*}{Methods} &
    \multirow{2}{*}{Size} &
    \multicolumn{3}{c}{Segmentation ($\mathcal{J}$\&$\mathcal{F}$)} &
    \multicolumn{1}{c}{Understanding (acc)} \\
    \cmidrule(lr){3-5} \cmidrule(lr){6-6}
     & & MeViS$_u$ & Ref-DAVIS17 & ReVOS & MVBench \\
    \midrule
    VideoLISA & 3.8B & 51.7 & 68.8 & -- & -- \\
    VISA      & 7B   & --   & 70.4 & 47.1 & -- \\
    Sa2VA     & 4B   & 52.1 & 73.8 & 53.2 & -- \\
    UniPixel  & 3B   & 59.7 & 74.2 & 62.1 & 62.5 \\
    UniPixel  & 7B   & 61.7 & 76.4 & 63.9 & 64.3 \\
    \midrule
    Ours & 3B & 61.8\inc{2.1} & 74.0\dec{0.2} & 62.7\inc{0.6} & 62.3\dec{0.2} \\
    Ours & 7B & \textbf{62.5}\inc{0.8} & \textbf{77.8}\inc{1.4} & \textbf{64.2}\inc{0.3} & \textbf{65.6}\inc{1.3} \\
    \bottomrule
  \end{tabular*}
  \label{tab:other_bench}
\end{table*}

We evaluate models on the E\mbox{-}VQA task using the ST\mbox{-}Evidence benchmark, from two categories: \textit{General Video LLMs} and \textit{Grounded Video LLMs}.

\noindent \textbf{General Video LLMs.} These models are designed for broad multimodal understanding and primarily output text. 
For closed-source models, we select three state-of-the-art multimodal LLMs: OpenAI-o3, Gemini-2.5-Flash, and Gemini-2.5-Pro as representatives. We also evaluate multiple recent open source MLLMs~\cite{damonlpsg2025videollama3, an2025llava, wang2025internvl3, bai2025qwen25vltechnicalreport,qwen2025qwen3vl}, with parameter counts ranging from 3B to 235B. 
Notably, while some general models (e.g., Gemini and Qwen3-VL) support image-level grounding by predicting bounding boxes or masks as strings, they cannot directly produce tracked video masks due to length and efficiency constraints.

\noindent \textbf{Grounded Video LLMs.} We evaluate two recent methods specialized for segmentation: Sa2VA~\cite{yuan2025sa2va} and UniPixel~\cite{liu2025unipixel}. These models equip the Multimodal LLM with SAM-2.1~\cite{ravi2024sam} as the segmentation head, enabling simultaneous text reasoning and pixel-level video mask prediction.

\subsection{Evaluation on ST-Evidence}
We conduct a comprehensive evaluation of baseline models on both variants of our benchmark. \textbf{ST-Evidence-Gen} requires models to predict temporal segments and object masks directly rather than selecting from options. Temporal segments are generated as text strings following the same format above. For spatial evidence, grounded Video LLMs can produce masks directly. However, since general Video LLMs lack native dense mask generation heads, we employ a two-step proxy evaluation for spatial evidence: the Video LLM first generates textual referring expressions for the evidence objects, which are then fed into a frozen video segmentation model (UniPixel-3B) to produce the final masks. To assess models' ability to reason jointly about answers and evidence, general Video LLMs are evaluated in a single inference pass and prompted to simultaneously output the answer, temporal segments, and spatial referring expressions. However, as current grounded LLMs often struggle with long, multi-task instructions and strict output formatting, we evaluate them using a sequential, multi-turn dialogue strategy in which the instructions and outputs from previous steps serve as context for subsequent subtasks.

On \textbf{ST-Evidence-MCQ}, we treat the three subtasks as independent multiple-choice questions. First, for answer prediction ($A$), the model is presented with the video, the question, and candidate answers, and must select the correct option. Second, for temporal evidence ($E_t$), we provide the video, the original question, and four candidate temporal segments formatted as text strings (e.g., ``A. [[$s_1, e_1$], \dots]''), querying the model to identify the time intervals that best support the answer. Finally, for spatial evidence ($E_s$), we construct the options as four distinct images, each visualizing a candidate mask overlaid with a red boundary on the target frame. The model receives the video, question, and these four visual options, and must select the one highlighting the correct objects that serve as the evidence.

\subsection{Experimental Results on ST-Evidence}
Evaluation results on ST-Evidence-Gen and ST-Evidence-MCQ are shown in Table~\ref{tab:gen_table} and Table~\ref{tab:mcq_table}, respectively.

\noindent \textbf{Performance on ST-Evidence-Gen.}
Table~\ref{tab:gen_table} details model performance across the three subtasks. Gemini-2.5-Pro leads in QA and Temporal Evidence ($E_t$), although its t-mean of 49.9 leaves substantial room for improvement. Among open-source General Video LLMs, the Qwen3-VL series exhibits substantial gains over Qwen2.5-VL. This improvement is likely attributable to its novel text-timestamp alignment mechanism, which adopts an interleaved ``timestamp-frame'' input format to enable fine-grained alignment between temporal information and visual content, thereby directly benefiting temporal evidence prediction. Consequently, Qwen3-VL-235B-A22B achieves the second-best performance on $E_t$.

Regarding Spatial Evidence ($E_s$), even the strongest General Video LLM obtains only 44.0 $\mathcal{J}$\&$\mathcal{F}$: Gemini-2.5-Pro ranks first at 44.0, followed by Gemini-2.5-Flash at 42.9, while Qwen3-VL-235B-A22B is the strongest open-source General Video LLM at 41.9. This underscores an inherent limitation in existing models: they often rely on language priors or shortcuts rather than grounding answers in specific visual evidence. Turning to Grounded Video LLMs, Sa2VA-Qwen2.5-VL-7B achieves a QA accuracy of 76.8\% but fails entirely on temporal segments (0.0 mIoU). Its low S-Evidence score further indicates a weakness in handling segmentation tasks that require complex reasoning rather than simple phrase grounding. Similarly, while UniPixel-3B produces temporal outputs, its poor performance reflects the lack of explicit temporal supervision during training. On S-Evidence, UniPixel-3B reaches 42.7 $\mathcal{J}$\&$\mathcal{F}$, outperforming all open-source General Video LLMs and OpenAI-o3, but remaining below Gemini-2.5-Pro and Gemini-2.5-Flash. Overall, these results highlight the significant challenges current models face in performing reliable spatio-temporal evidence-backed video question answering.

\noindent \textbf{Performance on ST-Evidence-MCQ.}
In the multiple-choice setting (Table~\ref{tab:mcq_table}), most General Video LLMs achieve competitive QA accuracies (approximately 68\%--83\%).
While all three closed-source models consistently perform near 80\%, Gemini-2.5-Pro attains the highest QA accuracy (82.82\%), while the open-source Qwen3-VL-235B-A22B is very competitive (81.59\%). However, a critical observation is that comparable QA performance does not imply similar grounding capabilities. For instance, while Video-LLaMA3-7B and LLaVA-OV-1.5-8B have similar QA scores (67.64\% and 69.03\%), they exhibit a substantial 17.18-point gap in temporal evidence accuracy. Regarding spatial evidence, OpenAI-o3 demonstrates significantly stronger visual grounding capabilities than Gemini-2.5-Flash, while Gemini-2.5-Pro is also competitive. Among open-source baselines, the Qwen-VL family—particularly the latest Qwen3-VL models—displays consistently strong performance in identifying evidence objects. In contrast, most other models perform near the random-guess baseline ($\sim$25\%), highlighting a widespread lack of fine-grained discriminative capability to distinguish between correct and incorrect spatial regions. Notably, Qwen3-VL-235B-A22B delivers the strongest overall open-source performance, achieving the best open-source QA and S-Evidence scores, while Qwen3-VL-30B-A3B attains the best open-source T-Evidence score (71.54). Gemini-2.5-Flash achieves the strongest overall T-Evidence performance (78.81).

\noindent \textbf{Influence of Model Scaling.}
We analyze the impact of model scaling using the Qwen2.5-VL and Qwen3-VL families on \textbf{ST-Evidence-Gen}. While increasing model size within a single family consistently boosts QA accuracy, improvements in Temporal and Spatial Evidence are marginal. Conversely, comparing across generations reveals significant gains. When comparing models of similar scale (e.g., Qwen2.5-VL-3B vs. Qwen3-VL-4B), the newer generation demonstrates clear improvements across all metrics. Most notably, the compact Qwen3-VL-4B matches or outperforms the massive Qwen2.5-VL-72B on every metric, with the most dramatic improvement observed in T-Evidence. This indicates that simple parameter scaling is insufficient to solve the E-VQA task. Instead, fundamental advances in model architecture, data quality, and training objectives are the primary drivers of performance improvement.

\subsection{A Baseline Model for the E-VQA Task}
To validate the effectiveness of our automated annotation pipeline and the quality of the proposed ST-Evidence-Instruct dataset, we train a baseline model building upon the UniPixel architecture. Specifically, we fine-tune the 3B and 7B variants of UniPixel, which integrate Qwen2.5-VL (3B/7B) as the base Multimodal LLM and SAM-2.1-Base+ as the mask decoder.

To prevent overfitting to the specific E-VQA task and preserve general capabilities, we incorporate existing video segmentation~\cite{seo2020urvos, pont20172017, yuan2025sa2va,ding2023mevis, athar2024vicas, yan2024visa, wang2023towards, deng2025motion} and general video understanding datasets~\cite{liu2024improved, maaz2024videogpt+} into the training mix, following the UniPixel training recipe. We employ LoRA~\cite{hu2022lora} to efficiently fine-tune the visual encoder and LLM, while the mask decoder is fully trained. The model is optimized to jointly predict the answer, temporal segments, and evidence object masks by minimizing a combination of next-token prediction loss (for QA and temporal segments) and mask decoding losses~\cite{ravi2024sam} (for spatial evidence).

The quantitative results are presented in the final rows of Table~\ref{tab:gen_table}. Compared to the original UniPixel baseline, our fine-tuned model yields slight gains in QA accuracy but achieves substantial improvements in both Temporal and Spatial Evidence. Notably, on T-Evidence, our 3B model achieves a competitive score of 26.9, outperforming the significantly larger QWen2.5-VL-72B. On S-Evidence, both our 3B and 7B models significantly outperform all other baselines. These results validate the effectiveness of our annotation pipeline and demonstrate the potential of our ST-Evidence-Instruct dataset to advance future research in spatio-temporal evidence grounding.

We further evaluate our model on several video segmentation and general video understanding benchmarks~\cite{ding2023mevis, pont20172017, li2024mvbench, yan2024visa}, comparing it with other grounded Video LLMs in Table~\ref{tab:other_bench}. Our 7B model consistently surpasses all baseline models on both segmentation and general understanding tasks. These results demonstrate that our model not only excels at the proposed E-VQA task but also retains robust pixel-level and semantic-level capabilities.

\subsection{Ablations and Deeper Analyses}

\textbf{Is the Annotated Evidence Truly Evidential?} We examine whether the annotated evidence object masks truly contain answer-supporting evidence, rather than merely salient but non-essential regions. We conduct experiments on a 200-sample subset of ST-Evidence-Gen using two representative general Video LLMs (Gemini-2.5-Flash and Qwen3-VL-8B). For each model, we evaluate three settings: (1) \textbf{Original}, with the unmodified video; (2) \textbf{Evidence Mask Corruption}, where annotated evidence regions are pixelated to make them visually unrecognizable; and (3) \textbf{Non-evidence Mask Corruption}, where randomly selected non-evidence regions are pixelated using the same procedure. If the annotations indeed capture evidential objects, QA performance should degrade substantially more under evidence corruption than under non-evidence corruption. As shown in Table~\ref{tab:obj_corruption}, corrupting evidence objects causes a large QA drop, while corrupting non-evidence objects yields only a minor drop for both models. This result supports the validity of our task formulation and the reliability of our evidence annotations.

\begin{table}[t]
  \caption{QA accuracy with corrupted regions on a subset of ST-Evidence-Gen. }
  \vspace{-5pt}
  \label{tab:obj_corruption}
  \centering
  \setlength{\tabcolsep}{4.0pt}
  \footnotesize
  \begin{tabular}{lccc}
    \toprule
    Model & Original & Evidence Corr. & Non-evidence Corr. \\
    \midrule
    Gemini-2.5-Flash & \textbf{81.32} & 61.12\dec{20.20} & 76.43\dec{4.89} \\
    Qwen3-VL-8B & \textbf{79.31} & 58.32\dec{20.99} & 73.66\dec{5.65} \\
    \bottomrule
  \end{tabular}
  \vspace{-10pt}
\end{table}

\textbf{Is the Proxy Segmenter an Evaluation Bottleneck?} A potential concern when evaluating General Video LLMs via a proxy segmenter is whether low spatial scores ($\mathcal{J}$\&$\mathcal{F}$) reflect the LLM's poor reasoning or simply the segmenter's inability to follow text descriptions. To isolate this, we choose 100 samples with manually annotated referring expressions and feed them into our proxy segmenter (UniPixel-3B), achieving a $\mathcal{J}$\&$\mathcal{F}$ score of 44.7, significantly outperforming the expressions generated by top models (e.g., Gemini-2.5-Pro at 25.6).
This massive performance gap confirms that the primary bottleneck in the E-VQA task mainly lies in the LLM's inability to ground to the correct evidence objects.

\textbf{Validation of the Automated Annotation Pipeline.} To empirically validate the quality of the evidence masks generated by our automated pipeline, we evaluated the pipeline's outputs directly against the human-verified ST-Evidence-Gen benchmark. The pipeline achieved a $\mathcal{J}$\&$\mathcal{F}$ score of 62.8, comfortably outperforming both our fine-tuned 7B model (54.1) and the 235B baseline (41.9). This verifies the effectiveness of our automatic annotation pipeline.

\subsection{Visualization}
\label{sec:vis}

\begin{figure}[tb]
    \centering
    \begin{overpic}[width=0.85\linewidth]{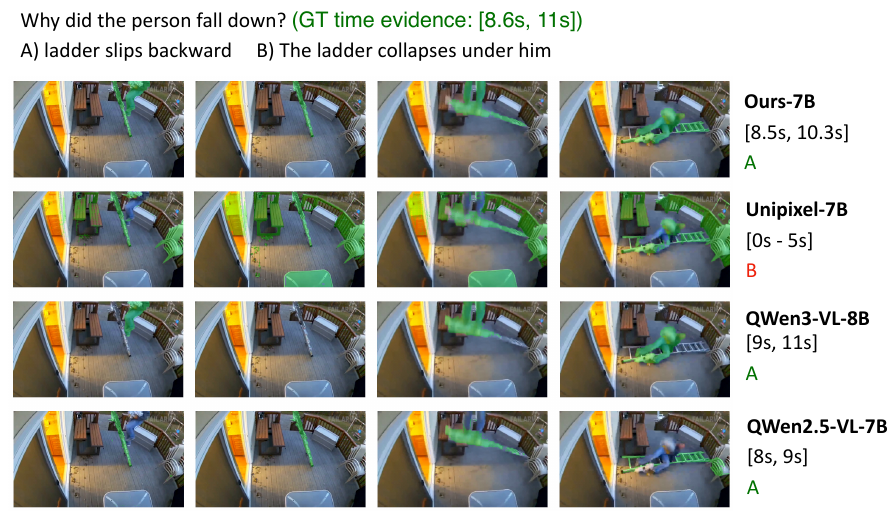}
        \put(83.10,32.20){\begingroup\setlength{\fboxsep}{0pt}\colorbox{white}{\rule{0pt}{2.7mm}\makebox[12.5mm]{}}\endgroup}
        \put(83.25,33.00){\scalebox{0.70}{\sffamily\bfseries UniPixel-7B}}
    \end{overpic}
    \vspace{-5pt}
    \captionof{figure}{\textbf{Comparison of models on ST-Evidence-Gen.} E-VQA requires models to jointly perform complex reasoning and provide dense spatial and temporal evidence.}
    \label{fig:compare}
    \vspace{-10pt}
\end{figure}

Figure~\ref{fig:compare} compares our model against baselines on ST-Evidence-Gen. To answer ``Why did the person fall down?'', a model needs to causally link the fall to the ladder slipping backward and localize the temporal interval in which both events occur. Furthermore, valid spatial evidence requires simultaneous, dense tracking of both entities. As illustrated, existing baseline models struggle to satisfy these strict evidence requirements: they either fail to capture the complete interaction between the person and the ladder (Qwen3-VL and Qwen2.5-VL) or ground unrelated background objects and output a highly inaccurate temporal window (UniPixel). In contrast, our model successfully segments both critical entities throughout the correct causal time span, demonstrating a clear synergy between high-level semantic reasoning and dense visual grounding. Additional qualitative results and failure-mode analyses are provided in the Supplementary Material.

\section{Conclusion}
We introduce Evidence-Backed Video Question Answering (E-VQA), a framework unifying question answering with dense spatio-temporal evidence grounding. We provide ST-Evidence, a human-verified benchmark, and ST-Evidence-Instruct, an instruction-tuning dataset that aligns answers with visual evidence. Fine-tuning on ST-Evidence-Instruct enhances both accuracy and grounding. Our work establishes a foundation for explainable and evidence-grounded Video LLMs that reason transparently over time and space. By requiring dense, video-aligned evidence alongside the answer, E-VQA bridges reasoning and perception, improves explainability and debuggability, and enables reliable use in applications that demand precision and transparency.

\bibliographystyle{splncs04}
\bibliography{main}

@inproceedings{xiao2024can,
  title={Can {I} trust your answer? visually grounded video question answering},
  author={Xiao, Junbin and Yao, Angela and Li, Yicong and Chua, Tat-Seng},
  booktitle={Proceedings of the IEEE/CVF Conference on Computer Vision and Pattern Recognition},
  pages={13204--13214},
  year={2024}
}

@misc{comanici2025gemini25pushingfrontier,
    title={Gemini 2.5: Pushing the Frontier with Advanced Reasoning, Multimodality, Long Context, and Next Generation Agentic Capabilities},
    author={Comanici, Gheorghe and others},
    year={2025},
    eprint={2507.06261},
    archivePrefix={arXiv},
    primaryClass={cs.CL},
    url={https://arxiv.org/abs/2507.06261},
    note={Accessed: 2026-06-26}
}

@article{yuan2025sa2va,
  title={{Sa2VA}: Marrying {SAM 2} with {LLaVA} for dense grounded understanding of images and videos},
  author={Yuan, Haobo and Li, Xiangtai and Zhang, Tao and Huang, Zilong and Xu, Shilin and Ji, Shunping and Tong, Yunhai and Qi, Lu and Feng, Jiashi and Yang, Ming-Hsuan},
  journal={arXiv preprint arXiv:2501.04001},
  year={2025}
}

@inproceedings{liu2025unipixel,
  title={{UniPixel}: Unified Object Referring and Segmentation for Pixel-Level Visual Reasoning},
  author={Liu, Ye and Ma, Zongyang and Pu, Junfu and Qi, Zhongang and Wu, Yang and Ying, Shan and Chen, Chang Wen},
  booktitle={Advances in Neural Information Processing Systems (NeurIPS)},
  year={2025}
}

@article{bai2024one,
  title={One token to seg them all: Language instructed reasoning segmentation in videos},
  author={Bai, Zechen and He, Tong and Mei, Haiyang and Wang, Pichao and Gao, Ziteng and Chen, Joya and Zhang, Zheng and Shou, Mike Zheng},
  journal={Advances in Neural Information Processing Systems},
  volume={37},
  pages={6833--6859},
  year={2024}
}

@misc{bai2025qwen25vltechnicalreport,
      title={{Qwen2.5-VL} Technical Report},
      author={Bai, Shuai and others},
      year={2025},
      eprint={2502.13923},
      archivePrefix={arXiv},
      primaryClass={cs.CV},
      url={https://arxiv.org/abs/2502.13923},
      note={Accessed: 2026-06-26}
}

@article{an2025llava,
  title={{LLaVA-OneVision-1.5}: Fully open framework for democratized multimodal training},
  author={An, Xiang and Xie, Yin and Yang, Kaicheng and Zhang, Wenkang and Zhao, Xiuwei and Cheng, Zheng and Wang, Yirui and Xu, Songcen and Chen, Changrui and Wu, Chunsheng and others},
  journal={arXiv preprint arXiv:2509.23661},
  year={2025}
}

@article{ravi2024sam,
  title={{SAM 2}: Segment anything in images and videos},
  author={Ravi, Nikhila and Gabeur, Valentin and Hu, Yuan-Ting and Hu, Ronghang and Ryali, Chaitanya and Ma, Tengyu and Khedr, Haitham and R{\"a}dle, Roman and Rolland, Chloe and Gustafson, Laura and others},
  journal={arXiv preprint arXiv:2408.00714},
  year={2024}
}

@article{patraucean2023perception,
  title={Perception test: A diagnostic benchmark for multimodal video models},
  author={Patraucean, Viorica and Smaira, Lucas and Gupta, Ankush and Recasens, Adria and Markeeva, Larisa and Banarse, Dylan and Koppula, Skanda and Malinowski, Mateusz and Yang, Yi and Doersch, Carl and others},
  journal={Advances in Neural Information Processing Systems},
  volume={36},
  pages={42748--42761},
  year={2023}
}

@inproceedings{CLEVRER2020ICLR,
  author    = {Kexin Yi and
               Chuang Gan and
               Yunzhu Li and
               Pushmeet Kohli and
               Jiajun Wu and
               Antonio Torralba and
               Joshua B. Tenenbaum},
  title     = {{CLEVRER:} Collision Events for Video Representation and Reasoning},
  booktitle = {ICLR},
  year      = {2020}
}

@inproceedings{wu2021star_situated_reasoning,
author = {Wu, Bo and Yu, Shoubin and Chen, Zhenfang and Tenenbaum, Joshua B and Gan, Chuang},
title = {STAR: A Benchmark for Situated Reasoning in Real-World Videos},
booktitle = {NeurIPS},
year = {2021}
}

@inproceedings{
hu2022lora,
title={Lo{RA}: Low-Rank Adaptation of Large Language Models},
author={Edward J Hu and Yelong Shen and Phillip Wallis and Zeyuan Allen-Zhu and Yuanzhi Li and Shean Wang and Lu Wang and Weizhu Chen},
booktitle={International Conference on Learning Representations},
year={2022}
}

@article{athar2024vicas,
author = {Athar, Ali and Deng, Xueqing and Chen, Liang-Chieh},
title = {{ViCaS}: A Dataset for Combining Holistic and Pixel-level Video Understanding using Captions with Grounded Segmentation},
journal = {CVPR},
year = {2025}
}

@inproceedings{xiao2021next,
  title={{NeXT-QA}: Next phase of question-answering to explaining temporal actions},
  author={Xiao, Junbin and Shang, Xindi and Yao, Angela and Chua, Tat-Seng},
  booktitle={Proceedings of the IEEE/CVF conference on computer vision and pattern recognition},
  pages={9777--9786},
  year={2021}
}

@article{wang2025internvl3,
  title={{InternVL3.5}: Advancing open-source multimodal models in versatility, reasoning, and efficiency},
  author={Wang, Weiyun and Gao, Zhangwei and Gu, Lixin and Pu, Hengjun and Cui, Long and Wei, Xingguang and Liu, Zhaoyang and Jing, Linglin and Ye, Shenglong and Shao, Jie and others},
  journal={arXiv preprint arXiv:2508.18265},
  year={2025}
}

@article{damonlpsg2025videollama3,
  title={{VideoLLaMA 3}: Frontier Multimodal Foundation Models for Image and Video Understanding},
  author={Zhang, Boqiang and others},
  journal={arXiv preprint arXiv:2501.13106},
  year={2025}
}

@misc{qwen2025qwen3vl,
  title   = {{Qwen3-VL}: Sharper Vision, Deeper Thought, Broader Action},
  author  = {{Qwen Team}},
  year    = {2025},
  month   = {sep},
  url     = {https://qwen.ai/blog?id=99f0335c4ad9ff6153e517418d48535ab6d8afef},
  note    = {Accessed: 2025-11-13}
}

@inproceedings{liu2024improved,
  title={Improved baselines with visual instruction tuning},
  author={Liu, Haotian and Li, Chunyuan and Li, Yuheng and Lee, Yong Jae},
  booktitle={Proceedings of the IEEE/CVF conference on computer vision and pattern recognition},
  pages={26296--26306},
  year={2024}
}

@article{maaz2024videogpt+,
  title={{VideoGPT+}: Integrating image and video encoders for enhanced video understanding},
  author={Maaz, Muhammad and Rasheed, Hanoona and Khan, Salman and Khan, Fahad},
  journal={arXiv preprint arXiv:2406.09418},
  year={2024}
}

@inproceedings{seo2020urvos,
  title={{URVOS}: Unified referring video object segmentation network with a large-scale benchmark},
  author={Seo, Seonguk and Lee, Joon-Young and Han, Bohyung},
  booktitle={European conference on computer vision},
  pages={208--223},
  year={2020},
  organization={Springer}
}

@article{pont20172017,
  title={The 2017 {DAVIS} challenge on video object segmentation},
  author={Pont-Tuset, Jordi and Perazzi, Federico and Caelles, Sergi and Arbel{\'a}ez, Pablo and Sorkine-Hornung, Alex and Van Gool, Luc},
  journal={arXiv preprint arXiv:1704.00675},
  year={2017}
}

@inproceedings{ding2023mevis,
  title={{MeViS}: A large-scale benchmark for video segmentation with motion expressions},
  author={Ding, Henghui and Liu, Chang and He, Shuting and Jiang, Xudong and Loy, Chen Change},
  booktitle={Proceedings of the IEEE/CVF international conference on computer vision},
  pages={2694--2703},
  year={2023}
}

@inproceedings{yan2024visa,
  title={{VISA}: Reasoning video object segmentation via large language models},
  author={Yan, Cilin and Wang, Haochen and Yan, Shilin and Jiang, Xiaolong and Hu, Yao and Kang, Guoliang and Xie, Weidi and Gavves, Efstratios},
  booktitle={European Conference on Computer Vision},
  pages={98--115},
  year={2024},
  organization={Springer}
}

@inproceedings{wang2023towards,
  title={Towards open-vocabulary video instance segmentation},
  author={Wang, Haochen and Yan, Cilin and Wang, Shuai and Jiang, Xiaolong and Tang, Xu and Hu, Yao and Xie, Weidi and Gavves, Efstratios},
  booktitle={proceedings of the IEEE/CVF international conference on computer vision},
  pages={4057--4066},
  year={2023}
}

@inproceedings{deng2025motion,
  title={Motion-grounded video reasoning: Understanding and perceiving motion at pixel level},
  author={Deng, Andong and Chen, Tongjia and Yu, Shoubin and Yang, Taojiannan and Spencer, Lincoln and Tian, Yapeng and Mian, Ajmal Saeed and Bansal, Mohit and Chen, Chen},
  booktitle={Proceedings of the Computer Vision and Pattern Recognition Conference},
  pages={8625--8636},
  year={2025}
}

@inproceedings{li2024mvbench,
  title={{MVBench}: A comprehensive multi-modal video understanding benchmark},
  author={Li, Kunchang and Wang, Yali and He, Yinan and Li, Yizhuo and Wang, Yi and Liu, Yi and Wang, Zun and Xu, Jilan and Chen, Guo and Luo, Ping and others},
  booktitle={Proceedings of the IEEE/CVF Conference on Computer Vision and Pattern Recognition},
  pages={22195--22206},
  year={2024}
}

@article{meng2025open,
  title={{Open-o3 Video}: Grounded Video Reasoning with Explicit Spatio-Temporal Evidence},
  author={Meng, Jiahao and Li, Xiangtai and Wang, Haochen and Tan, Yue and Zhang, Tao and Kong, Lingdong and Tong, Yunhai and Wang, Anran and Teng, Zhiyang and Wang, Yujing and others},
  journal={arXiv preprint arXiv:2510.20579},
  year={2025}
}

@article{wei2022chain,
  title={Chain-of-thought prompting elicits reasoning in large language models},
  author={Wei, Jason and Wang, Xuezhi and Schuurmans, Dale and Bosma, Maarten and Xia, Fei and Chi, Ed and Le, Quoc V and Zhou, Denny and others},
  journal={Advances in neural information processing systems},
  volume={35},
  pages={24824--24837},
  year={2022}
}

@inproceedings{huang2024vtimellm,
  title={{VTimeLLM}: Empower {LLM} to grasp video moments},
  author={Huang, Bin and Wang, Xin and Chen, Hong and Song, Zihan and Zhu, Wenwu},
  booktitle={Proceedings of the IEEE/CVF Conference on Computer Vision and Pattern Recognition},
  pages={14271--14280},
  year={2024}
}

@article{qian2024momentor,
  title={Momentor: Advancing video large language model with fine-grained temporal reasoning},
  author={Qian, Long and Li, Juncheng and Wu, Yu and Ye, Yaobo and Fei, Hao and Chua, Tat-Seng and Zhuang, Yueting and Tang, Siliang},
  journal={arXiv preprint arXiv:2402.11435},
  year={2024}
}

@article{guo2024vtg,
  title={{VTG-LLM}: Integrating Timestamp Knowledge into Video {LLMs} for Enhanced Video Temporal Grounding},
  author={Guo, Yongxin and Liu, Jingyu and Li, Mingda and Tang, Xiaoying and Chen, Xi and Zhao, Bo},
  journal={arXiv preprint arXiv:2405.13382},
  year={2024}
}

@article{munasinghe2024videoglamm,
  title={{VideoGLaMM}: A Large Multimodal Model for Pixel-Level Visual Grounding in Videos},
  author={Shehan Munasinghe and Hanan Gani and Wenqi Zhu and Jiale Cao and Eric Xing and Fahad Khan and Salman Khan},
  journal={ArXiv},
  year={2024},
  url={https://arxiv.org/abs/2411.04923},
  note={Accessed: 2026-06-26}
}

@article{feng2025video,
  title={{Video-R1}: Reinforcing video reasoning in {MLLMs}},
  author={Feng, Kaituo and Gong, Kaixiong and Li, Bohao and Guo, Zonghao and Wang, Yibing and Peng, Tianshuo and Wu, Junfei and Zhang, Xiaoying and Wang, Benyou and Yue, Xiangyu},
  journal={arXiv preprint arXiv:2503.21776},
  year={2025}
}

@article{li2025videochatr1,
  title={{VideoChat-R1}: Enhancing Spatio-Temporal Perception via Reinforcement Fine-Tuning},
  author={Li, Xinhao and Yan, Ziang and Meng, Desen and Dong, Lu and Zeng, Xiangyu and He, Yinan and Wang, Yali and Qiao, Yu and Wang, Yi and Wang, Limin},
  journal={arXiv preprint arXiv:2504.06958},
  year={2025}
}

@inproceedings{wang2025video,
  title={{Video-RTS}: Rethinking reinforcement learning and test-time scaling for efficient and enhanced video reasoning},
  author={Wang, Ziyang and Yoon, Jaehong and Yu, Shoubin and Islam, Md Mohaiminul and Bertasius, Gedas and Bansal, Mohit},
  booktitle={Proceedings of the 2025 Conference on Empirical Methods in Natural Language Processing},
  pages={28114--28128},
  year={2025}
}

@misc{OpenAI_o3_2024,
  author       = {{OpenAI}},
  title        = {{OpenAI o3} Model},
  year         = {2024},
  howpublished = {\url{https://platform.openai.com/docs/models/o3}},
  note         = {Accessed: 2025-11-14}
}

@misc{openai_gpt4o,
  author       = {{OpenAI}},
  title        = {{GPT-4o} System Card},
  year         = {2024},
  publisher    = {arXiv},
  version      = {arXiv:2410.21276v1},
  url          = {https://arxiv.org/abs/2410.21276},
  note         = {Accessed: 2026-06-26}
}

@article{you2025seg,
  title={{Seg-R1}: Segmentation Can Be Surprisingly Simple with Reinforcement Learning},
  author={You, Zuyao and Wu, Zuxuan},
  journal={arXiv preprint arXiv:2506.22624},
  year={2025}
}

@article{gong2025reinforcing,
  title={Reinforcing video reasoning segmentation to think before it segments},
  author={Gong, Sitong and Zhang, Lu and Zhuge, Yunzhi and Jia, Xu and Zhang, Pingping and Lu, Huchuan},
  journal={arXiv preprint arXiv:2508.11538},
  year={2025}
}

@article{jin2025interrvos,
  title={{InterRVOS}: Interaction-aware Referring Video Object Segmentation},
  author={Jin, Woojeong and Kim, Seongchan and Lee, Jaeho and Kim, Seungryong},
  journal={arXiv preprint arXiv:2506.02356},
  year={2025}
}

@article{wang2025time,
  title={{Time-R1}: Post-Training Large Vision Language Model for Temporal Video Grounding},
  author={Wang, Ye and Wang, Ziheng and Xu, Boshen and Du, Yang and Lin, Kejun and Xiao, Zihan and Yue, Zihao and Ju, Jianzhong and Zhang, Liang and Yang, Dingyi and others},
  journal={arXiv preprint arXiv:2503.13377},
  year={2025}
}

@inproceedings{liang2025fine,
  title={Fine-grained Spatiotemporal Grounding on Egocentric Videos},
  author={Liang, Shuo and Zhong, Yiwu and Hu, Zi-Yuan and Tao, Yeyao and Wang, Liwei},
  booktitle={Proceedings of the IEEE/CVF International Conference on Computer Vision},
  pages={9385--9395},
  year={2025}
}

@inproceedings{lin2025glus,
  title={{GLUS}: Global-local reasoning unified into a single large language model for video segmentation},
  author={Lin, Lang and Yu, Xueyang and Pang, Ziqi and Wang, Yu-Xiong},
  booktitle={Proceedings of the Computer Vision and Pattern Recognition Conference},
  pages={8658--8667},
  year={2025}
}

@inproceedings{han2025videoespresso,
  title={{VideoEspresso}: A large-scale chain-of-thought dataset for fine-grained video reasoning via core frame selection},
  author={Han, Songhao and Huang, Wei and Shi, Hairong and Zhuo, Le and Su, Xiu and Zhang, Shifeng and Zhou, Xu and Qi, Xiaojuan and Liao, Yue and Liu, Si},
  booktitle={Proceedings of the Computer Vision and Pattern Recognition Conference},
  pages={26181--26191},
  year={2025}
}

@article{sun2025sama,
  title={{SAMA}: Towards Multi-Turn Referential Grounded Video Chat with Large Language Models},
  author={Sun, Ye and Zhang, Hao and Ding, Henghui and Zhang, Tiehua and Ma, Xingjun and Jiang, Yu-Gang},
  journal={arXiv preprint arXiv:2505.18812},
  year={2025}
}

@inproceedings{yuan2025videorefer,
  title={{VideoRefer} suite: Advancing spatial-temporal object understanding with video {LLM}},
  author={Yuan, Yuqian and Zhang, Hang and Li, Wentong and Cheng, Zesen and Zhang, Boqiang and Li, Long and Li, Xin and Zhao, Deli and Zhang, Wenqiao and Zhuang, Yueting and others},
  booktitle={Proceedings of the Computer Vision and Pattern Recognition Conference},
  pages={18970--18980},
  year={2025}
}

@inproceedings{zhou2025strefer,
  title={{STRefer}: Empowering Video {LLMs} with Space-Time Referring and Reasoning via Synthetic Instruction Data},
  author={Zhou, Honglu and Peng, Xiangyu and Kendre, Shrikant and Ryoo, Michael S and Savarese, Silvio and Xiong, Caiming and Niebles, Juan Carlos},
  booktitle={Proceedings of the IEEE/CVF International Conference on Computer Vision},
  pages={4289--4300},
  year={2025}
}

@article{wang2025vgr,
  title={{VGR}: Visual grounded reasoning},
  author={Wang, Jiacong and Kang, Zijian and Wang, Haochen and Jiang, Haiyong and Li, Jiawen and Wu, Bohong and Wang, Ya and Ran, Jiao and Liang, Xiao and Feng, Chao and others},
  journal={arXiv preprint arXiv:2506.11991},
  year={2025}
}

@inproceedings{wu2025number,
  title={Number it: Temporal grounding videos like flipping manga},
  author={Wu, Yongliang and Hu, Xinting and Sun, Yuyang and Zhou, Yizhou and Zhu, Wenbo and Rao, Fengyun and Schiele, Bernt and Yang, Xu},
  booktitle={Proceedings of the Computer Vision and Pattern Recognition Conference},
  pages={13754--13765},
  year={2025}
}

@inproceedings{lu2025vited,
  title={{VITED}: Video temporal evidence distillation},
  author={Lu, Yujie and Song, Yale and Wang, William and Torresani, Lorenzo and Nagarajan, Tushar},
  booktitle={Proceedings of the Computer Vision and Pattern Recognition Conference},
  pages={8501--8511},
  year={2025}
}

@article{chen2024cg,
  title={{CG-Bench}: Clue-grounded question answering benchmark for long video understanding},
  author={Chen, Guo and Liu, Yicheng and Huang, Yifei and He, Yuping and Pei, Baoqi and Xu, Jilan and Wang, Yali and Lu, Tong and Wang, Limin},
  journal={arXiv preprint arXiv:2412.12075},
  year={2024}
}

@article{ahmad2025videomolmo,
  title={{VideoMolmo}: Spatio-Temporal Grounding Meets Pointing},
  author={Ahmad, Ghazi Shazan and Heakl, Ahmed and Gani, Hanan and Shaker, Abdelrahman and Shen, Zhiqiang and Khan, Fahad Shahbaz and Khan, Salman},
  journal={arXiv preprint arXiv:2506.05336},
  year={2025}
}

@misc{cheng2025vstarbenchmarkingvideollmsvideo,
      title={{V-STaR}: Benchmarking {Video-LLMs} on Video Spatio-Temporal Reasoning},
      author={Zixu Cheng and Jian Hu and Ziquan Liu and Chenyang Si and Wei Li and Shaogang Gong},
      year={2025},
      eprint={2503.11495},
      archivePrefix={arXiv},
      primaryClass={cs.CV},
      url={https://arxiv.org/abs/2503.11495},
      note={Accessed: 2026-06-26}
}

@misc{carion2025sam3segmentconcepts,
      title={{SAM 3}: Segment Anything with Concepts},
      author={Nicolas Carion and Laura Gustafson and Yuan-Ting Hu and Shoubhik Debnath and Ronghang Hu and Didac Suris and Chaitanya Ryali and Kalyan Vasudev Alwala and Haitham Khedr and Andrew Huang and Jie Lei and Tengyu Ma and Baishan Guo and Arpit Kalla and Markus Marks and Joseph Greer and Meng Wang and Peize Sun and Roman Rädle and Triantafyllos Afouras and Effrosyni Mavroudi and Katherine Xu and Tsung-Han Wu and Yu Zhou and Liliane Momeni and Rishi Hazra and Shuangrui Ding and Sagar Vaze and Francois Porcher and Feng Li and Siyuan Li and Aishwarya Kamath and Ho Kei Cheng and Piotr Dollár and Nikhila Ravi and Kate Saenko and Pengchuan Zhang and Christoph Feichtenhofer},
      year={2025},
      eprint={2511.16719},
      archivePrefix={arXiv},
      primaryClass={cs.CV},
      url={https://arxiv.org/abs/2511.16719},
      note={Accessed: 2026-06-26}
}

@inproceedings{wang2024stair,
  title={{STAIR}: Spatial-Temporal Reasoning with Auditable Intermediate Results for Video Question Answering},
  author={Wang, Yueqian and Wang, Yuxuan and Chen, Kai and Zhao, Dongyan},
  booktitle={Proceedings of the AAAI Conference on Artificial Intelligence},
  volume={38(17)},
  pages={19215--19223},
  year={2024},
  doi={10.1609/aaai.v38i17.29890}
}

@inproceedings{li2023transtr,
  title={Discovering Spatio-Temporal Rationales for Video Question Answering},
  author={Li, Yicong and Xiao, Junbin and Feng, Chun and Wang, Xiang and Chua, Tat-Seng},
  booktitle={Proceedings of the IEEE/CVF International Conference on Computer Vision},
  pages={13869--13878},
  year={2023}
}

@inproceedings{lei2020tvqaplus,
  title={{TVQA+}: Spatio-Temporal Grounding for Video Question Answering},
  author={Lei, Jie and Yu, Licheng and Berg, Tamara and Bansal, Mohit},
  booktitle={Proceedings of the 58th Annual Meeting of the Association for Computational Linguistics},
  pages={8211--8225},
  year={2020},
  doi={10.18653/v1/2020.acl-main.730}
}

@inproceedings{grauman2022ego4d,
  title={{Ego4D}: Around the World in 3,000 Hours of Egocentric Video},
  author={Grauman, Kristen and others},
  booktitle={Proceedings of the IEEE/CVF Conference on Computer Vision and Pattern Recognition},
  pages={18995--19012},
  year={2022}
}

\end{document}